\def\year{2021}\relax
\documentclass[letterpaper]{article} 

\usepackage{aaai20}  
\usepackage{times}  
\usepackage{helvet} 
\usepackage{courier}  
\usepackage[hyphens]{url}  
\usepackage{graphicx} 
\usepackage{graphicx}  
\usepackage{epsfig}
\usepackage{url}

\usepackage{makecell}
\usepackage{mathtools}
\usepackage{comment}

\usepackage{xcolor}

\usepackage{amsmath,amssymb} 

\usepackage{appendix}
\usepackage{multirow}
\usepackage{bm}

\usepackage{xcolor}

\newcommand{\Eq}[1]{Eq.~(\ref{eq:#1})}
\newcommand{\eq}[1]{\Eq{#1}}

\newcommand{\fig}[1]{Fig.~\ref{fig:#1}}

\newcommand{\tabl}[1]{Table~\ref{table:#1}}

\usepackage{graphicx}
\usepackage{amsmath}
\usepackage{amssymb}
\usepackage{booktabs}

\newcommand{\ie}{\textit{i}.\textit{e}.}
\newcommand{\eg}{\textit{e}.\textit{g}.}
\usepackage{xspace}

\makeatletter
\newif\if@restonecol
\makeatother

\usepackage[linesnumbered,ruled,vlined]{algorithm2e}
\usepackage{algpseudocode}
\usepackage{amsmath}

\title{Deep Unsupervised Image Hashing  by Maximizing Bit Entropy}
\author{ \\ \Large \textbf{
    Yunqiang Li and
Jan van Gemert} 
\\
Vision Lab, Delft University of
Technology, Netherlands
\\  \{y.li-19, \ j.c.vangemert\}@tudelft.nl
\\ 
}

\begin{document}

\maketitle

\begin{abstract}
Unsupervised hashing is important for indexing huge image or video collections without having expensive annotations available. Hashing aims to learn short binary codes for compact storage and efficient semantic retrieval.
We propose an unsupervised deep hashing layer called Bi-half Net that maximizes entropy of the binary codes. Entropy is maximal when both possible values of the bit are uniformly (half-half) distributed. To maximize bit entropy, we do not add a term to the loss function as this is difficult to optimize and tune. Instead, we design a new parameter-free network layer to explicitly force continuous image features to approximate the optimal half-half bit distribution. This layer is shown to minimize a penalized term of the Wasserstein distance between the learned continuous image features and the optimal half-half bit distribution. Experimental results on the image datasets Flickr25k, Nus-wide, Cifar-10, Mscoco, Mnist and the video datasets Ucf-101 and Hmdb-51
 show that our approach leads to compact codes and compares favorably to the current state-of-the-art.
\end{abstract}

\section{Introduction}

Semantically similar images or videos can be found by comparing their output features in the last layer of a deep network. Such features are typically around 1,000 continuous floating point values~\cite{He2016Deep}, which is already too slow and large for moderately sized datasets of a few million samples. Speed and storage are greatly improved by replacing the continuous features with just a small number of bits. Unsupervised hashing aims to learn compact binary codes that preserves semantic similarity without making use of any annotated label supervision and is thus of great practical importance for indexing huge visual collections.


\begin{figure}[t] \vspace{-0.1in}
	\centering
	\includegraphics[width= 0.47\textwidth]{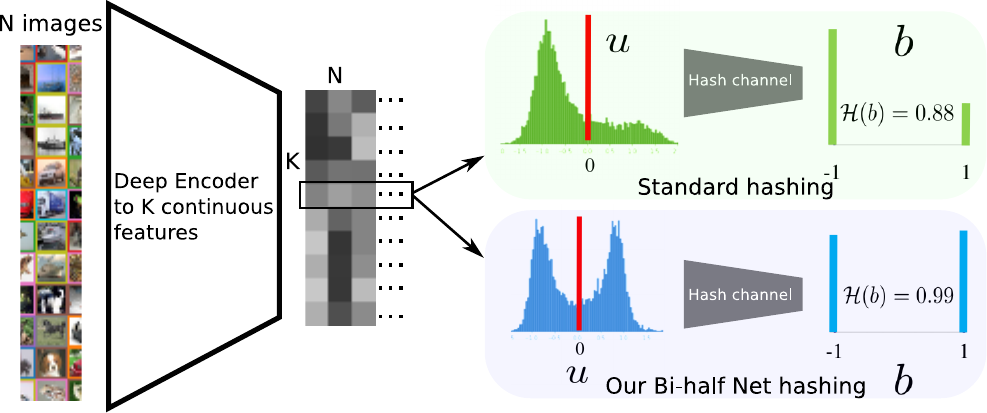}
	\vspace{-0.1in}  \caption{Hashing compresses $N$ images to $K$ bits per image. $K$ continuous features are learned, and thresholded to $K$ binary values.
		We see the continuous to binary transition as a lossy communication channel --a hash channel-- between a single continuous value $u$ to a single discrete binary value $b$. The green histograms (standard hashing) and blue histograms (our approach) show how a single feature is distributed over the $N$ images. Instead of adding an additional loss term, we design a Bi-half layer to explicitly maximizes the bit capacity in the hash channel, leading to more informative hash codes, as measured by the entropy $\mathcal{H}$ of the bits over the images which leads to improved hashing accuracy.
\vspace{-0.15in}
}
	\label{fig:distribution}
\end{figure}





In this paper, as illustrated in \fig{distribution}, we see the transition from a continuous variable  to a discrete binary variable as a lossy communication channel. The capacity of a hash bit as measured by the entropy is maximized when it is half-half distributed: Half of the images are encoded with $-1$ and the other half of the images is encoded with $+1$. We minimize the information loss in the hash channel by forcing the continuous variable to be half-half distributed. Other methods have optimized entropy by adding an additional term to the loss~\cite{erin2015deep,Liu2014Discrete,shen2018unsupervised,Weiss2008Spectral,xu2013harmonious} which adds an additional hyper-parameter to tune and is difficult to optimize. Instead, we propose Bi-half: A new parameter-free network layer which is shown to minimize the optimal transport cost as measured by the Wasserstein distance.  We here explicitly design a new layer to maximize the bit capacity in the hash channel, leading to compact and informative hash codes which yield excellent hashing similarity accuracy.

We have the following contributions. \vspace{-0.06in}
\begin{itemize}
\item A simple, parameter-free, bi-half layer to maximize hash channel information capacity; \vspace{-0.06in}
\item A Wasserstein distance is minimized end-to-end to align continuous features with the optimal discrete distribution;\vspace{-0.06in}
\item We study 2 alternatives to maximizing bit entropy using an additional term in the loss; \vspace{-0.06in}  
\item We show state-of-the-art results for unsupervised hashing on 5 image datasets, and 2 video datasets, and make our code available\footnote{\scriptsize{https://github.com/liyunqianggyn/Deep-Unsupervised-Image-Hashing}}.
\end{itemize}

\section{Related Work}

\textbf{Amount of supervision.}
Hashing methods can be grouped into data-independent hashing methods and data-dependent hashing methods.
Data-independent hashing methods~\cite{Datar04localitysensitivehashing,LSH1,Kulis09kernelizedlocalitysensitive,Kulis2009Fast,Mu2010Non,Raginsky2009Locality} design hash function independent of a dataset. In contrast, data-dependent hashing methods can exploit the data distribution. As such, with the availability of  labeled training data,
supervised hashing methods~\cite{Chang2012Supervised,LaiPLY15,Lin2014Fast,Raziperchikolaei2015Optimizing,Shen2015Supervised,Zhang2014Supervised} learn hash codes by optimizing class labels. Particularly successful supervised image hashing methods use deep learning~\cite{CaoCVPR18,cao2017hashnet,LaiPLY15,Li2017NIPS,li2019push,Liu2016DeepSH,Liu17cvpr,yuan2018relaxation,chen2018deep} to learn feature representations
and binary codes. Supervised methods work well, yet rely on data annotations done by humans, which are expensive  or difficult to obtain.
Unsupervised hashing methods~\cite{Gong11iterativequantization,He2013K,Jiang2015Scalable,Kong2012Isotropic,Liu2014Discrete,liu2011hashing,Weiss2008Spectral} skip this problem, as they do not rely on annotation labels. Recent unsupervised hashing methods rely on deep learning for representation learning~\cite{dai2017stochastic,yang2019distillhash,shen2018unsupervised,ghasedi2018unsupervised,lin2016learning,erin2015deep}.
We follow these works and focus on the unsupervised setting.

\noindent \textbf{Quantization from continuous to discrete values.} The typical approach for deep learning hashing is to optimize a continuous output and in the last step quantize the continuous values to discrete values. The current approach~\cite{chen2018learning,lu2017deep,NIPS_greedy_hash}  is to apply  a $\mathrm{sign}$ function, where all negative values are set to $-1$ and all positive values are set to $+1$. We argue that the $\mathrm{sign}$ function  is not information efficient.
For example, we set the continuous features of one  dimension for 4 images to be $[0.2,0.8,1.5,3]$. Passing them through the $\mathrm{sign}$ function will binarize them all to same value $+1$ and thus the bit has no discriminative information for these 4 images. In this paper we focus on this loss of information and learn to discretize based on maximum bit capacity over images.

\noindent \textbf{Obtaining gradients for binary codes.}
A major challenge of learning hash codes with deep nets is that the desired discrete hash output codes have no continuous derivatives and cannot be directly optimized by gradient descent.
By  the continuous relaxation~\cite{CaoCVPR18,Jiang_2017_CVPR,Liu2016DeepSH,Zhao_2015_CVPR}, a continuous space is optimized instead and the continuous  values are quantized to binary codes.
Such methods are approximations as they do not optimize the binary codes directly.
The continuation based hashing methods~\cite{cao2017hashnet,LaiCVPR2015}  gradually approximate the non-smooth \emph{sign} function with \emph{sigmoid} or \emph{tanh},
but unfortunately comes with the drawback that such relaxation inevitably becomes more non-smooth during training which slows down convergence, making it difficult to optimize.
To overcome these problems, a recent simple and efficient method called greedy hash~\cite{NIPS_greedy_hash},  uses the $\mathrm{sign}$ function in the forward pass to
directly optimize binary codes. The optimization is done with the straight-through estimator~\cite{bengio2013estimating} which after quantization computes gradients by simply ignoring the quantization function during training. This optimization is simple and works well in practice.
Yet, it ignores bit information capacity and thus may lead to redundant codes. In this work we use the same straight-through estimator to obtain gradients for binary codes while focusing on maximizing bit information capacity to obtain compact and discriminative hash codes.

%

\noindent \textbf{Information theory in hashing.}
Many popular unsupervised feature learning methods~\cite{belghazi2018mine,chen2016infogan,Jolliffe2002Principal} are based on  information theory to find good features. In hash learning, some methods~\cite{erin2015deep,Liu2014Discrete,shen2018unsupervised,Weiss2008Spectral,xu2013harmonious} proposed to add an additional term in the loss function to encourage each bit to have a 50\% chance of being one or zero, to maximize bit entropy. It is, however, difficult to balance the added loss term with other terms in the loss, which requires careful hyper-parameter tuning of how much to weight each term in the loss.
 Instead of adding an additional loss term and an additional hyper-parameter, we design a new network layer without any additional
parameters to explicitly force continuous image features to approximate the optimal half-half bit distribution.
Some non-deep learning approaches~\cite{jegou2008hamming,zhang2010self} directly threshold the learned feature vectors at their median point, which have shown excellent performance.
Yet, it is a suboptimal solution under deep learning scenario since the median point should be dynamically adapted to random sample statistic computed over each minibatch.
We are inspired by their works, and aim to generalize such ideas to an end-to-end deep learnable setting. We cast it into an optimal transport problem and directly quantize the continuous features into half-half distributed binary codes by minimizing the Wasserstein distance between the continuous distribution and a prior half-half distribution.

\section{Approach} \label{approach}
This paper we maximize the hash channel capacity to design a parameter-free bi-half coding layer.
We will first introduce some notations.
Let $\mathbf{X} = \{ \mathbf{x_i}\}_{i = 1}^N$ denote $N$ training images.
The images are encoded to $K$ compact binary codes $\mathbf{B}\in \{1,-1\}^{N\times K}$, which also denotes the output of our hash coding layer.
 $\mathbf{U} \in \mathbb{R}^{N\times K}$ would be
expressed as the continuous feature representations in the last layer of a standard neural network, \eg,  an encoder, which serves as the input to our hash coding layer.

\subsection{Maximizing hash channel capacity}


We see the transition from a continuous variable $U$ to a binary code variable $B$ as a lossy communication channel.
Per channel, the \emph{maximum} transmitted information from continuous variable ${U}$  to binary variable ${B}$,   known as channel capacity~\cite{cover2012elements,shannon1948mathematical}, is:
 \begin{equation}
{C} = \max_{p(u)} {I}({U}; {B}),
\label{eq:Channel_capacity}
\end{equation}
where the maximum is taken over all possible input distributions $p(u)$
and ${I}({U}; {B})$ denotes mutual information between variable ${U}$  and binary variable ${B}$. We aim to maximize the channel capacity. To maximize channel capacity ${C}$ we first rewrite the mutual information term ${I}({U}; {B})$ in \eq{Channel_capacity} in terms of entropy:
\begin{equation}
{I}({U}; {B}) = \mathcal{H}({B}) -
\mathcal{H}({B}|{U}),
\label{eq:ChanCapAsEntropy}
\end{equation}
where $\mathcal{H}({B})$ and  $\mathcal{H}({B}|{U})$ denote entropy and  conditional entropy respectively.
Thus, maximizing channel capacity ${C}$ in \eq{Channel_capacity} is equivalent to  maximizing the entropy $\mathcal{H}({B})$ of ${B}$ and minimizing the conditional entropy $\mathcal{H}({B}|{U})$.

The entropy $\mathcal{H}({B})$ in \eq{ChanCapAsEntropy} should be maximized. Since ${B}$ is a discrete binary variable, its  entropy is maximized when it is half-half distributed: \vspace{0.02in}
    \begin{equation} \vspace{0.02in}
p({B} = +1) = p({B} = -1) = \frac{1}{2}.
\label{eq:entropy}
\end{equation}

The conditional entropy $\mathcal{H}({B}|{U})$ in \eq{ChanCapAsEntropy} should be minimized.
Give a certain continuous value ${u}$, the transmission probability $p_{{u}}(\text{pos})$ is defined as how probable a $+1$ binary output value is and the transmission probability $p_{{u}}(\text{neg})$ is defined as how probable the binary value $-1$ is. These are probabilities and thus are non-negative and sum to one as $p_{{u}}(\text{pos}) + p_{{u}}(\text{neg}) = 1$ and $0 \leq p_{{u}}(\text{pos}), p_{{u}}(\text{neg}) \leq 1$. Then the conditional entropy is computed as:
\begin{gather}\vspace{0.03in}
\begin{aligned}
 \mathcal{H}({B}|{U}) =  & \int_{u \in \mathcal{U}}  p(u)
   \mathcal{H}({B}|{U} = u) du \\
 =
&-  \int_{u \in \mathcal{U}}  p(u)\Big(p_{{u}}(\text{pos})\log p_{{u}}(\text{pos})  \\
& \ \ \ \ \  \ \   + p_{{u}}(\text{neg}) \log p_{{u}}(\text{neg}) \Big) du,
\end{aligned}
\label{eq:conditional_entropy}
\raisetag{20pt}
\end{gather}
%
%
which is between 0 and 1,  and \Eq{conditional_entropy} is thus minimized for setting the $\mathcal{H}({B}|{U} = u)$ to 0, \ie:  $-  \int_{u \in \mathcal{U}}  p(u) 0 \  du = 0$. This minimum is obtained when either  $p_{{u}}(\text{pos}) = 1$ or $ p_{{u}}(\text{neg}) = 1$, which means that there is no stochasticity for a certain  continuous value $u$, and its binary value is deterministically transmitted.


In the following, we maximize the entropy of binary variables
by encouraging the continuous feature distribution $p(u)$ to align with the ideal half-half distributed distribution $p(b)$ in \eq{entropy}. To minimize \eq{conditional_entropy}, we first start with a non-deterministic transmission probability during training, but since we train to align $p(u)$ with the half-half distribution of $+1$ and $-1$, this allows us at test time to simply use the $\mathrm{sign}$ function as  a deterministic function for quantization to  guarantee minimizing \eq{conditional_entropy}.

 \subsection{Bi-half layer for quantization}



To align the continuous feature distribution with the ideal prior half-half distributed distribution from \eq{entropy} we use  Optimal Transport (OT)~\cite{villani2003topics}.
 Optimal Transport aims to find a minimal cost plan for moving one unit of mass from one location  $\mathbf{x}$ to one other location $\mathbf{y}$ between two  probability distributions $\mathbb{P}_r$ and $\mathbb{P}_g$.
 When  $\mathbb{P}_r$ and $\mathbb{P}_g$  are only accessible through discrete
samples, the corresponding optimal transport cost can be defined as:
  \begin{equation}
\pi_0= \mathop{\min}_{\pi \in \Pi(\mathbb{P}_r, \mathbb{P}_g)} 	\left \langle \pi, \mathbf{D}\right \rangle_F,
\label{rewasserp0}
\end{equation}
where $\Pi(\mathbb{P}_r, \mathbb{P}_g)$ is the space of joint probability measures with marginals $\mathbb{P}_r $ and $\mathbb{P}_g$,
and $\pi$ is the general probabilistic coupling that indicates how much mass is transported to push distribution $\mathbb{P}_r$  towards  distribution $\mathbb{P}_g$.
 	The $\left \langle ., .\right \rangle_F$ denotes the Frobenius dot product,  and $\mathbf{D} \geq 0$ is the cost function matrix whose element  ${D}(i, j) =  d(\mathbf{x}, \mathbf{y})$ denotes the  non-negative cost to move a probability mass from location $\mathbf{x}$ to location $\mathbf{y}$.
 When the cost is defined as a distance, OT is referred to as a Wasserstein distance. Specifically, if  $d(\mathbf{x}, \mathbf{y})$ is the squared Euclidean distance, it is the Earth mover's distance, which is also known as the 1-Wasserstein distance. We optimize the 1-Wasserstein distance because it is flexible and easy to bound.

With a randomly sampled mini-batch of $M$ samples,  the corresponding empirical distributions  of  the continuous variable $U$ and binary variable $B$, $P_u$ and $P_b$, can be written as:
  \begin{equation}
P_u = \sum_{i=1}^M p_i\delta_{u_i}, \ \ \  P_b = \sum_{j=1}^2 q_j\delta_{b_j},
\label{rewasserp}
\end{equation}
 where $\delta_\mathbf{x}$ is the Dirac function at location $\mathbf{x}$.  The $p_i$  and $q_j$ are
  the probability mass associated to the corresponding location  $u_i$ and ${b_j}$, where the total mass is one, \ie: $\sum_{i=1}^M p_i = 1$ and  $\sum_{j=1}^2 q_j = 1$. Particularly,  a binary variable only has two locations $b_1$ and $b_2$, with the corresponding mass $q_1$ and $q_2$.

    For the ideal prior half-half distribution in \eq{entropy}, the probability mass $q_1$  at location $b_1$ is equal to the probability mass $q_2$  at location $b_2$ that is $q_1 = q_2 = \frac{1}{2}$.
  The  hash coding strategy is to find the optimal transport coupling $\pi_0$  by minimizing the
 1-Wasserstein distance $W_1(P_u, P_b)$:
\begin{equation}
 \pi_0 = \mathop{\min}_{\pi \in \Pi({P}_u, {P}_b)}\  \ 	\sum_i\sum_j \pi_{ij}(u_i - b_j)^2,
\label{1_wesserstan_distance}
\end{equation}
where $\Pi({P}_u, {P}_b)$ is the set of all joint probability distributions $\pi_{ij}$, \emph{i.e.} all probabilistic couplings,
with marginals ${P}_u$ and ${P}_b$, respectively.

\begin{figure}[t] 
  \centering
  \includegraphics[width=0.49\textwidth]{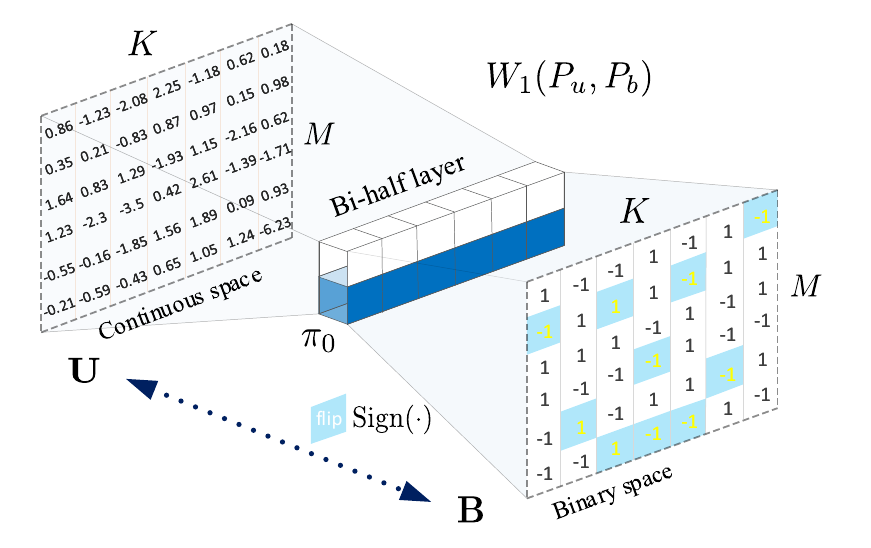}
  \caption{The proposed bi-half layer.
    $M$ is the mini-batch size and $K$ is the feature dimensions.
    A bi-half layer (middle part in white and blue) is used to quantize continuous features in $\mathbf{U}$ into binary codes in $\mathbf{B}$ via minimizing $W_1(P_u, P_b)$ in Eq.(\ref{1_wesserstan_distance}). The assignment strategy is the optimal probabilistic coupling $\pi_0$.
     For each bit, \ie per column of $\mathbf{U}$, we first rank its elements and
      then the top half elements is assigned to $+1$ and the remaining half elements to $-1$.   In contrast,  the commonly used sign function directly during training assigns the continuous
       features to their nearest binary codes which minimizes the Euclidean distance. The blue boxes indicate where our method differs from the sign function as the code in that position should flip.
   }
  \label{bi_half_layer}
\end{figure}

By optimizing Eq. (\ref{1_wesserstan_distance}), we  find an optimal transport plan $\pi_0 \in \Pi({P}_u, {P}_b)$ for one hash bit to quantize the encoded features into half-half distributed binary codes.
 For a single hash bit in $M$ samples, with a continuous feature vector $\mathbf{u} \in \mathbb{R}^M$,  we first simply sort the elements of $\mathbf{u}$ over all mini-batch images, and then assign the top half elements of sorted $\mathbf{u}$ to $+1$ and assign the remaining elements to $-1$, that is:
 \begin{equation}
\mathbf{b}= \mathrm{\pi_0}\ (\mathbf{u}) =
\begin{cases}
+1,  & \mathrm{ top \ half \ of \ sorted}\ \mathbf{u}\\
-1, & \mathrm{otherwise}
\end{cases}.
\label{behalf}
\end{equation}
We implement above equation as a new simple hash coding layer, dubbed \textbf{\emph{bi-half layer}} shown in Fig.\ref{bi_half_layer}, to quantize the continuous feature into half-half distributed binary code for each hash channel. The proposed bi-half layer can be easily embedded into current deep architectures
 to automatically generate higher quality binary codes.
 During training,  the  transmission stochasticity introduced by random small batches as shown  in \eq
 {conditional_entropy} can also improve the model generalization capability as the same effect of denoising.

\vspace{0.5in}

\textbf{Optimization.}
 The discrete binary codes $\mathbf{B}$ have no continuous derivatives and
cannot be directly optimized by gradient descent.
Fortunately, some recent works on binarized neural networks (BNNs) have explored to use a proxy derivative approximated by straight through estimator (STE)~\cite{bengio2013estimating}
 to avoid the vanishing gradients.
We  use the same straight-through estimator to obtain the gradients. Specifically, we expect $\mathbf{U}$ and $\mathbf{B}$ have the same update states in backward pass to match the forward goal.

  Given time-step $t$, the current states  are denoted as  $\mathbf{U}_t$ and $\mathbf{B}_t$. In time-step $t+1$, we force their update states to be same that   $\mathbf{U}_{t+1} = \mathbf{B}_{t+1}$. Considering the simplest SGD algorithm, we have $\mathbf{U}_{t+1} = \mathbf{U}_{t} - lr * \frac {\partial \mathcal{L}} {\partial \mathbf{U}_{t}}$ and  $\mathbf{B}_{t+1} = \mathbf{B}_{t} - lr * \frac {\partial \mathcal{L}} {\partial \mathbf{B}_{t}}$ with learning rate $lr$ and loss function $\mathcal{L}$ where  $\mathcal{L}$ can be any loss function you need to use, \eg \ reconstruction loss, cross entropy loss and so on, then the gradient of $\mathbf{U}_{t}$ is computed as
  $\frac {\partial \mathcal{L}} {\partial \mathbf{U}_{t}} = \frac {\partial \mathcal{L}} {\partial \mathbf{B}_{t}} + \gamma(\mathbf{U}_{t} - \mathbf{B}_{t})$ with  $\gamma = \frac{1}{lr}$. The forward pass and backward pass are concluded as:
 \begin{equation}
	\begin{split}
\text{Forward:} & \ \ \ \mathbf{B} = \mathrm{\mathbf{\pi}_0}\ (\mathbf{U}),
 \\
\text{Backward:} & \ \ \   \frac {\partial \mathcal{L}} {\partial \mathbf{U}} = \frac {\partial \mathcal{L}} {\partial \mathbf{B}} + \gamma(\mathbf{U} - \mathbf{B}).
	\end{split}
\label{eq:gradienet}
\end{equation}
In forward pass, the continuous feature is optimally quantized to half-half distributed binary codes.
In backward pass,  the proposed proxy derivative can automatically encourage the continuous feature
distribution to align with the ideal half-half distributed distribution.

\section{Experiments}

\begin{figure*}[t] \vspace{-0.2in}
\centering
\begin{tabular}{llllllll}
\multicolumn{4}{c}{
\includegraphics[width=0.19\textwidth]{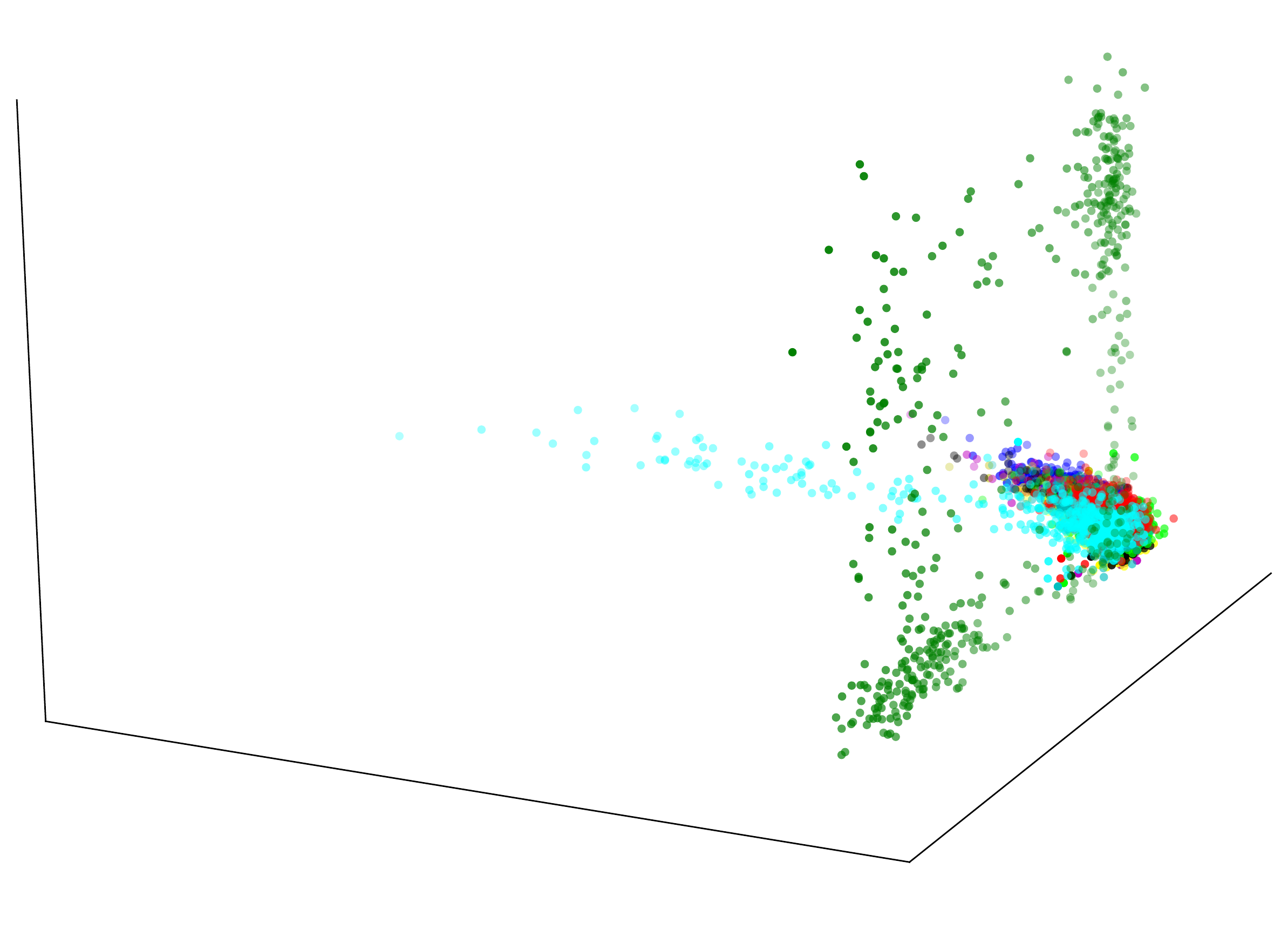} \ \   \  \
\includegraphics[width=0.19\textwidth]{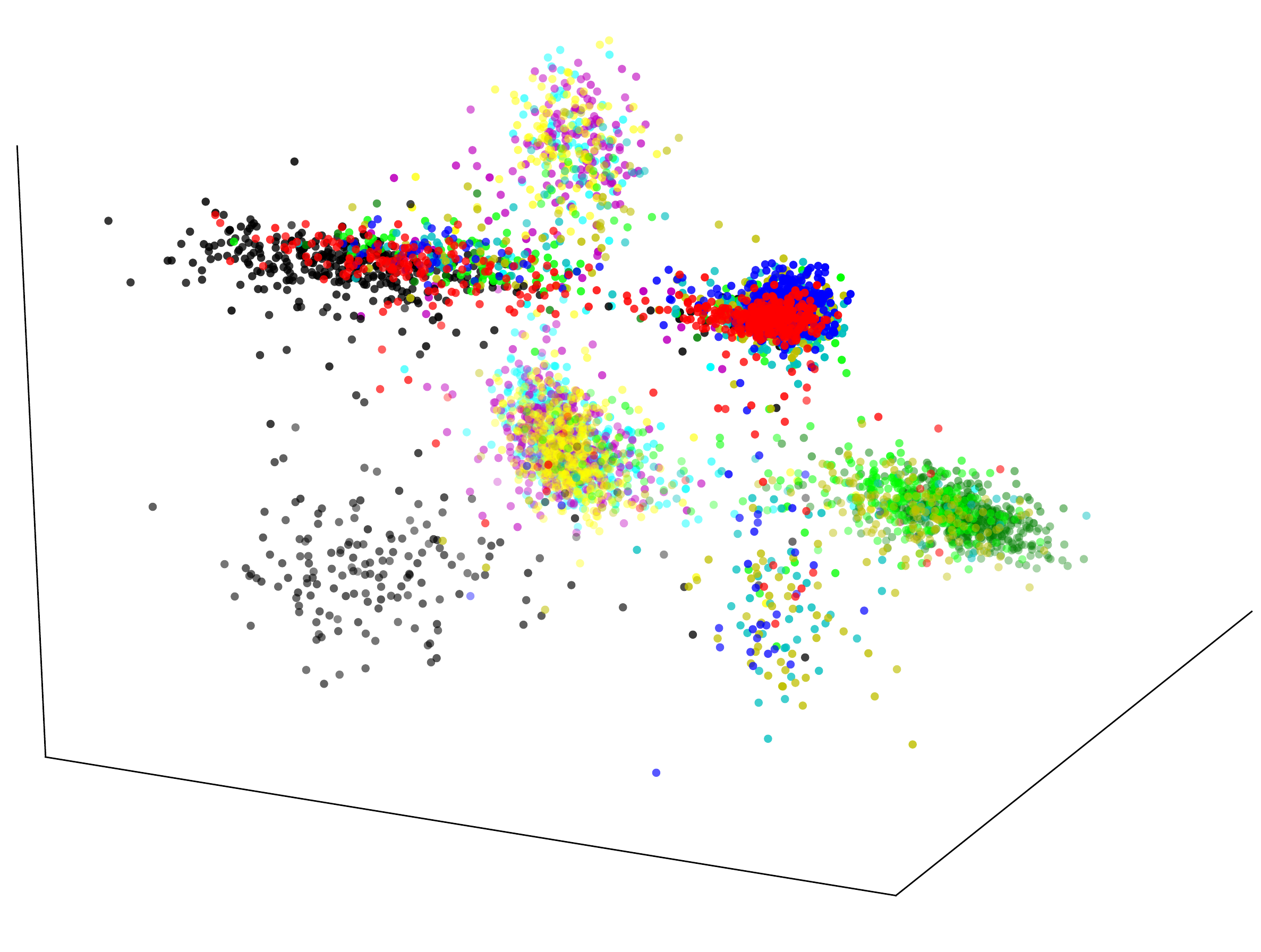}  \ \   \ \
\includegraphics[width=0.192\textwidth]{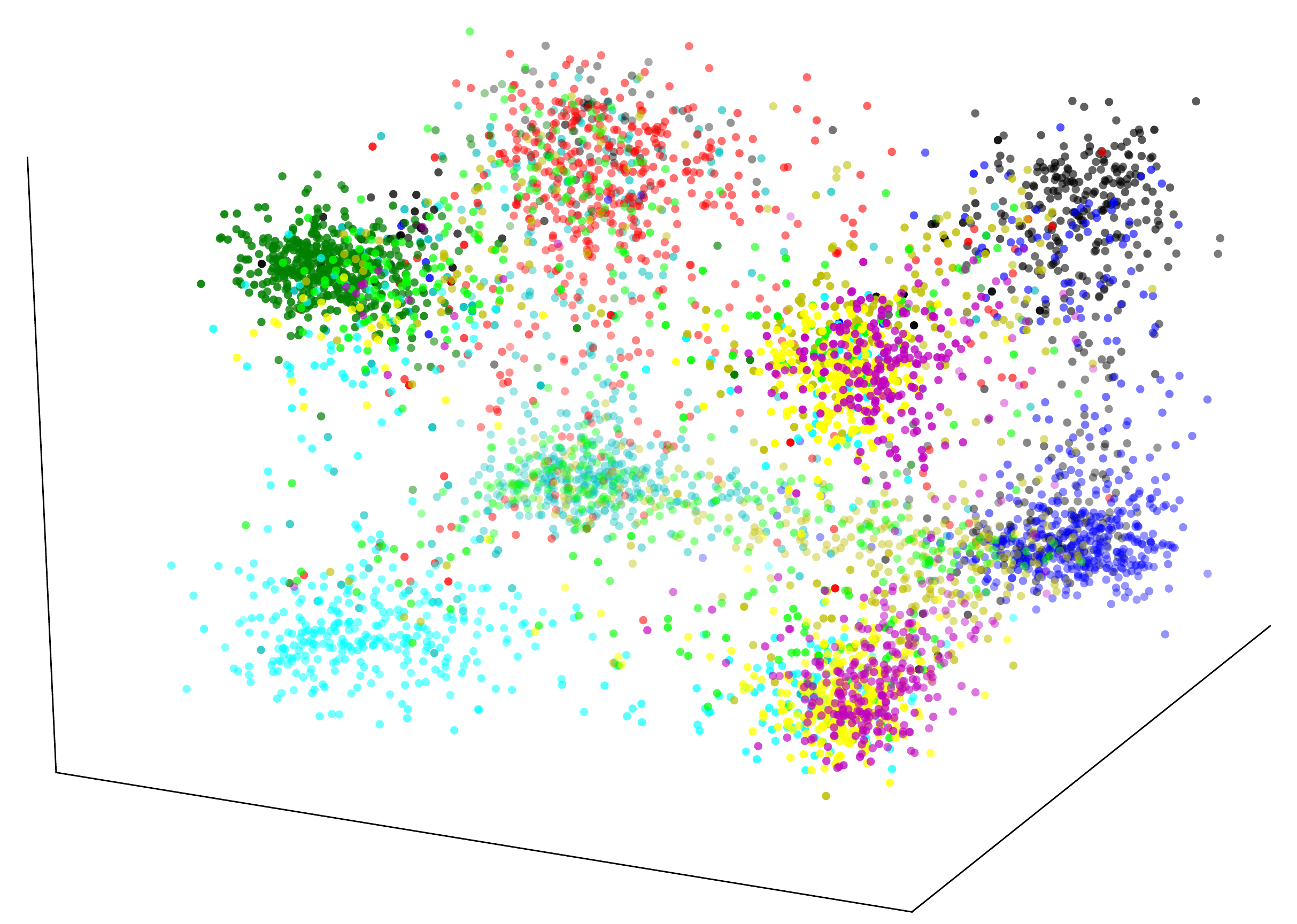} \ \
\includegraphics[width=0.1\textwidth]{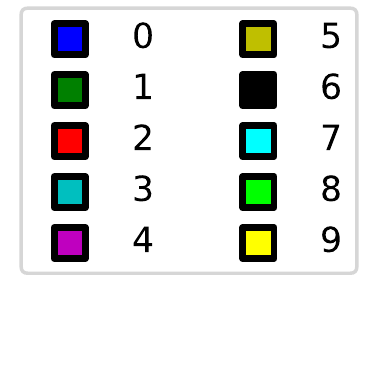}
}\\

\multicolumn{3}{c}{
\ \ \ \ \ \ \ \ \ \ \ \ \ \ \  \ \ \ \ \ \ \ \ \ \ \ \ \  \ \ \ \ \    (a) Sign Layer  \ \ \ \ \ \ \  \ \ \ \ \ \  \ \ \ \ \ \ \ \ (b) Sign + Reg \ \  \ \ \ \ \ \ \ \ \  \ \  \ \ \ \   (c)  Bi-half Layer
 }
 \vspace{-0.08in}
  \\
  \\
\multicolumn{4}{c}{
\includegraphics[width=0.186\textwidth]{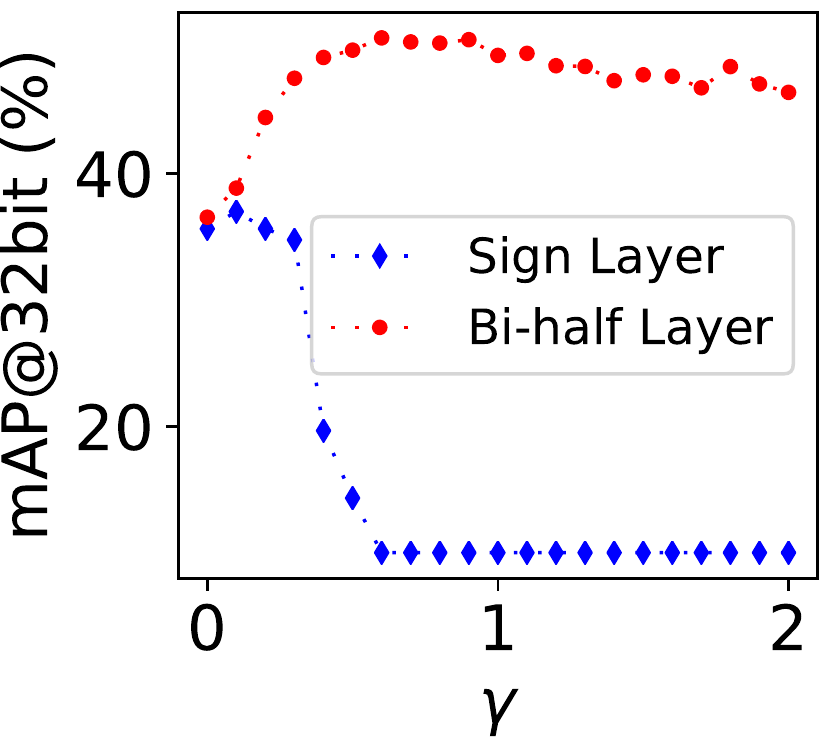}
\includegraphics[width=0.206\textwidth]{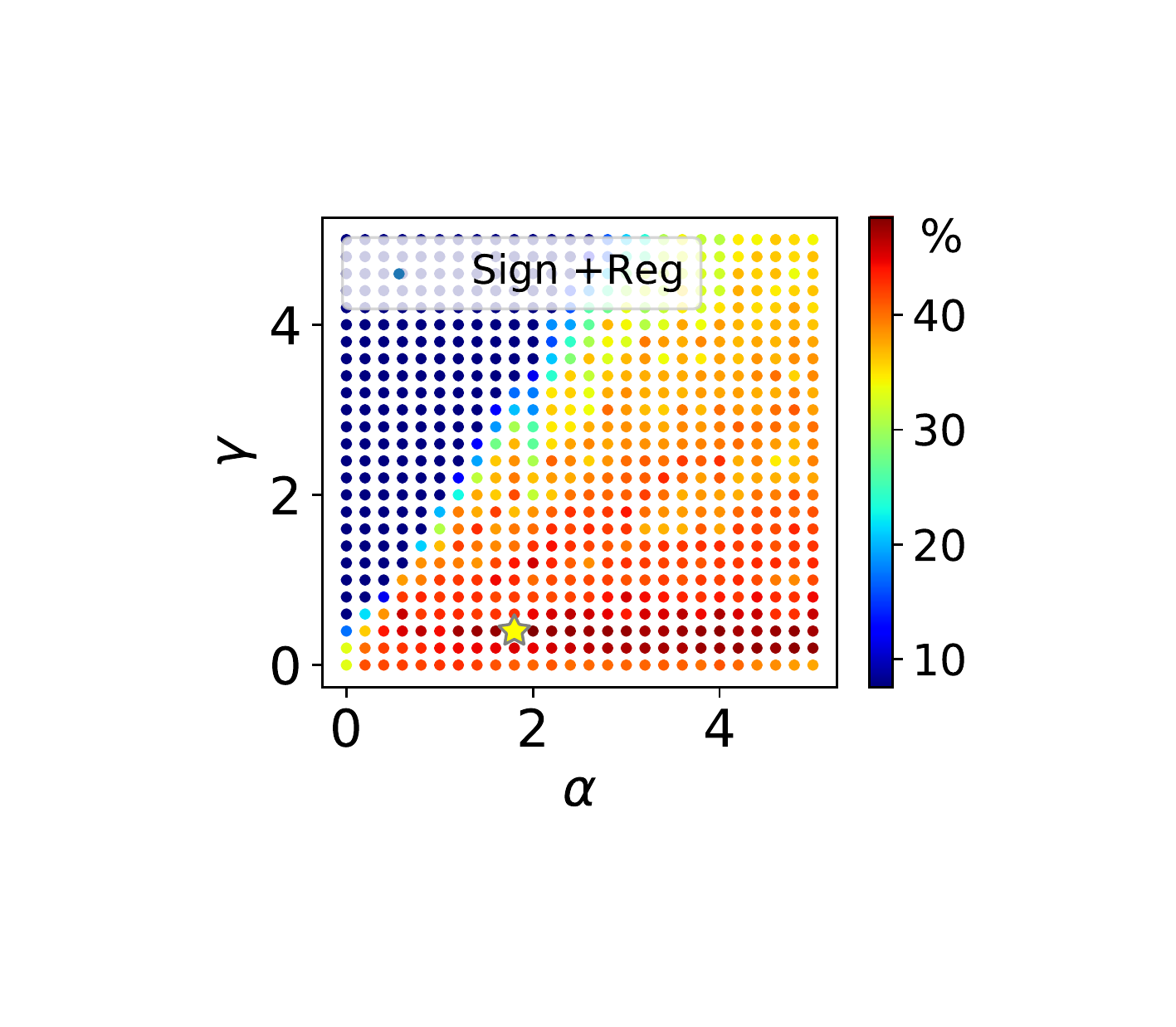}
\includegraphics[width=0.186\textwidth]{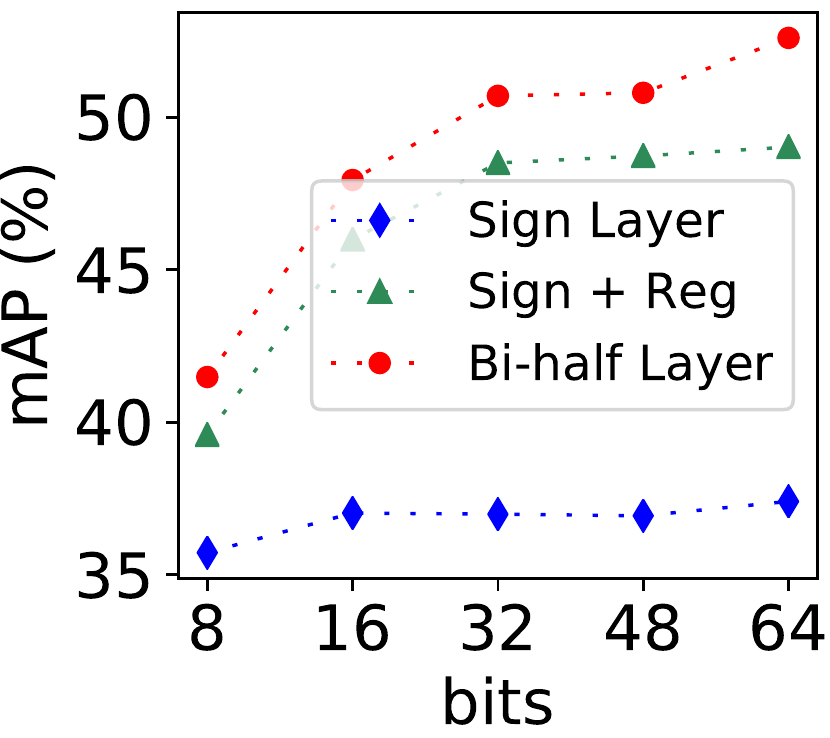} \ \
\includegraphics[width=0.2\textwidth]{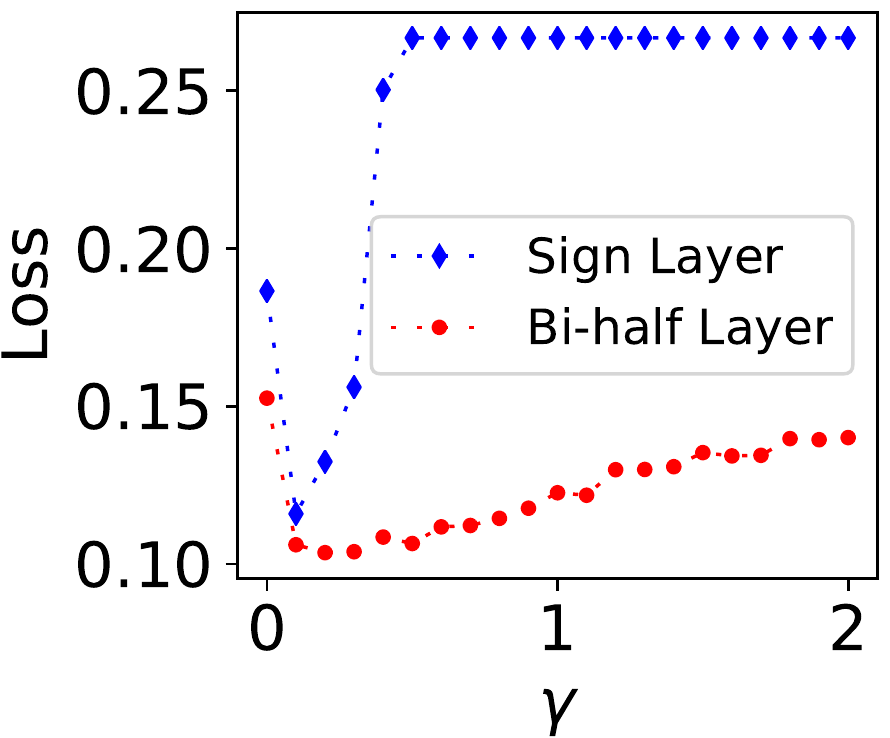}
\includegraphics[width=0.206\textwidth]{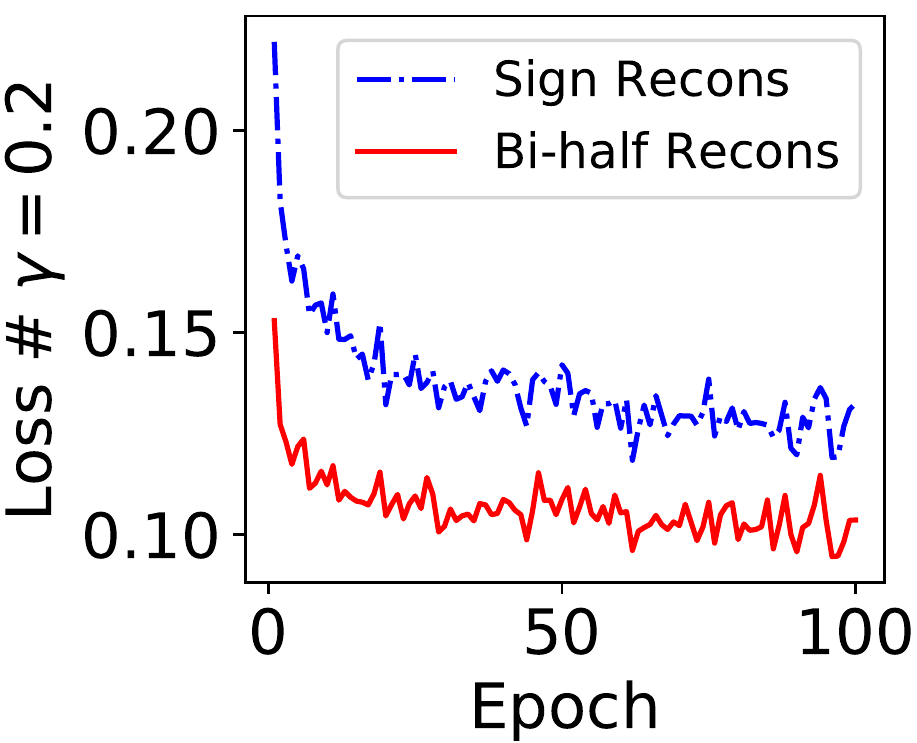}
}	
   \\
   \multicolumn{4}{l}{
   \ \ \ \ \ \ \ \ \ \ \ \ \ \ \ \ \ \ \ \ \  \ \ \ \ \ \ \  \ \ \ \ \ \ \ \ \ \ \ \ \ \ \ \
   (d) Retrieval Results
     \ \ \ \ \ \ \ \ \ \ \ \ \ \ \ \ \ \ \ \ \ \  \ \ \ \ \ \ \ \ \ \ \ \ \ \ \ \ \ \ \ \  \ \ \ \ \ \ \ \  \ \ \ \ \ \ \ \  \  \ \ \ \ \ \ \ \  \ \ \
   (e) Reconstruction Loss
   }
   \vspace{-0.1in}
\end{tabular}

\caption{  Train an AutoEncoder from scratch on Mnist dataset.  The top row (a, b, c) visualizes the continuous feature distributions
 before binarization over different methods by training the network with 3 hash bits.
 (d) shows the corresponding retrieval results.
 We compare bi-half layer with sign layer and sign+reg. In specific,  sign+reg uses an additional entropy regularization term to optimize entropy, while it is hard to balance the added term.
(e) shows the reconstruction loss curves for sign layer and bi-half layer.
Generating informative binary codes in latent space can help to do reconstruction.
 } \vspace{-0.15in}
 \label{fig:AE}
\end{figure*}

%

\textbf{Datasets.}
$\bullet$ \emph{Flickr25k}~\cite{huiskes2008mir} contains 25k images categorized into 24 classes. Each image is annotated with at least one label. Following~\cite{yang2019distillhash}, 2,000 random images are queries and from the remaining images 5,000 random images are training set.

$\bullet$  \emph{Nus-wide}~\cite{chua2009nus} has around 270k images with 81 classes. To fairly compare with other methods, we consider two versions. Nus-wide(I), following~\cite{shen2018unsupervised}, uses the 21 most frequent classes for evaluation. Per class, 100 random images form the query set and the remaining images form the retrieval database and training set. Nus-wide(II), following ~\cite{yang2019distillhash}, uses the 10 most popular classes where 5,000 random images form the test set, and the remaining images are the retrieval set.  From the retrieval set,  10,500 images are randomly selected  as the training set.

$\bullet$ \emph{Cifar-10}~\cite{krizhevsky2009learning} consists of $60$k  color images  categorized into 10 classes. 
In the literature there are also two experimental settings.
In Cifar-10(I), following~\cite{NIPS_greedy_hash}, 1k images per class (10k images in total) form the test query set, and the remaining 50k images are used  for training.  For Cifar-10(II), following~\cite{yang2019distillhash}, randomly selects 1,000 images per class as queries and 500 as training images, and the retrieval set has all  images except for the
query set.

$\bullet$  \emph{Mscoco}~\cite{lin2014microsoft} is a dataset for multiple tasks. We
use the pruned set as~\cite{cao2017hashnet} with 12,2218 images from
80 categories. We randomly select 5,000 images as queries
with the rest used as database, from which 10,000
images are chosen for training.

$\bullet$ \emph{Mnist}~\cite{lecun1998gradient}   contains 70k gray-scale $28 \times 28$ images of hand written digits from ``0" to ``9" across 10 classes. 1,000 images per
class are randomly selected as queries  and the remaining images as training set and database.

$\bullet$ \emph{Ucf-101}~\cite{soomro2012ucf101} contains 13,320 action instances from 101 human action classes. All the videos are downloaded from YouTube.
 The average duration per video is about 7 seconds.

$\bullet$ \emph{Hmdb-51}~\cite{kuehne2011hmdb} includes 6,766 videos from 51 human action
categories.  The average duration of each video is about 3 seconds.
   For both Ucf-101 and Hmdb-51 datasets,  we use the provided split~1, where per class 30\% of the videos are used for testing and  the rest 70\%  for training and retrieval.

\noindent \textbf{Implementation details.}
Our code is available online\footnote{\scriptsize{https://github.com/liyunqianggyn/Deep-Unsupervised-Image-Hashing}}. \\
$\bullet$ \emph{Image setup.}
For Mnist image dataset, we train an AutoEncoder from scratch. The details will be described in  the corresponding subsection.
For other image datasets,
 an ImageNet pre-trained VGG-16~\cite{simonyan2014very} is used as our backbone where following~\cite{NIPS_greedy_hash,shen2018unsupervised,yang2019distillhash} an additional fc layer is used for dimensionality reduction. Our bi-half layer is appended to generate the binary codes.
During training, we use Stochastic Gradient Descent(SGD) as the optimizer with   a momentum of $0.9$ and a weight decay of $5\times 10^{-4}$ and a batch size of $32$.
In all experiments, the initial learning rate is set as $0.0001$ and we divide the learning rate by $10$ when the loss stop decreasing. The hyper-parameters $\gamma$ is tuned by cross-validation on training set and set as   $\gamma = 3 \times \frac{1}{N \cdot K}$.

$\bullet$ \emph{Video setup.}
Two 3D CNNs pre-trained on kinetics~\cite{kay2017kinetics}, ResNet-34~\cite{hara2017learning} and ResNet-101~\cite{hara2018can}, are used as backbones where we append bi-half layer to replace the  last fc layer.
Following the setting of~\cite{hara2018can}, we use SGD as optimizer with a momentum of 0.9 and a
weight decay of 0.001.  The learning
rate starts from 0.1, and is divided by 10 after the validation loss saturates.

\noindent \textbf{Evaluation metrics.}
We adopt semantic
similarity (by labels) as evaluation ground truth, which is widely used in the unsupervised hashing literature,
for instance, AGH~\cite{liu2011hashing}, SADH~\cite{shen2018unsupervised} and DeepBit~\cite{lin2016learning}.
Specifically, for multi-label datasets Flickr25k, Nus-wide and Mscoco, the true
neighbors are defined based on whether two images share at least one common label. We measure the performance of compared methods based on the standard evaluation metrics:
 Mean Average Precision (mAP), Precision-Recall curves (PR) and TopN-precision curves with top $N$ returned samples. In our experiments, $N$ is set to 5,000. 


\begin{figure*}[t] \vspace{-0.05in}
\centering
\begin{tabular}{@{}c@{}c@{}c@{}@{}c@{}c@{}c@{}}
\includegraphics[width=0.1945\textwidth]{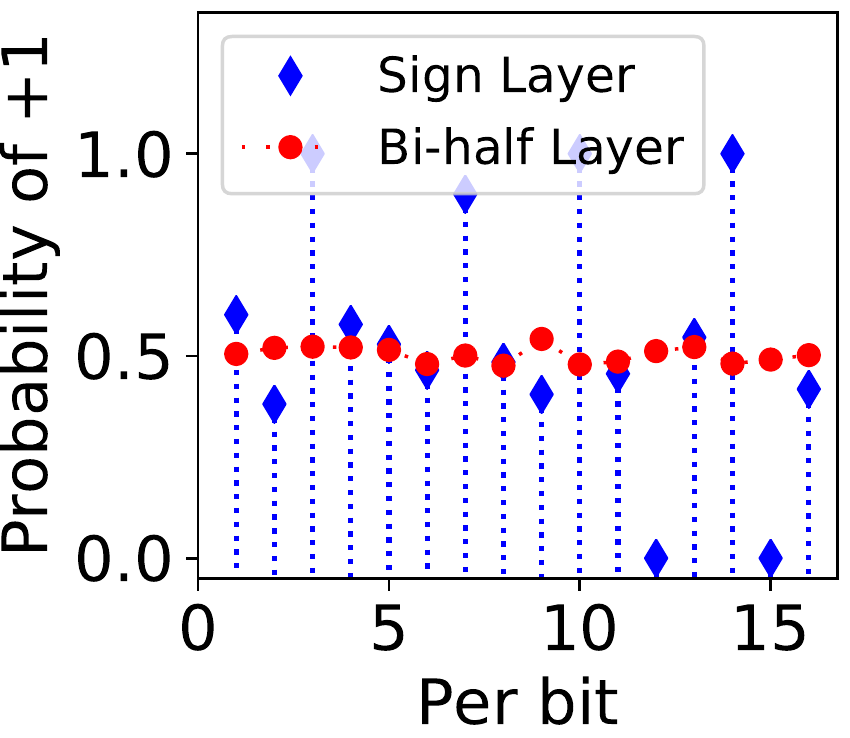}&
\includegraphics[width=0.19\textwidth]{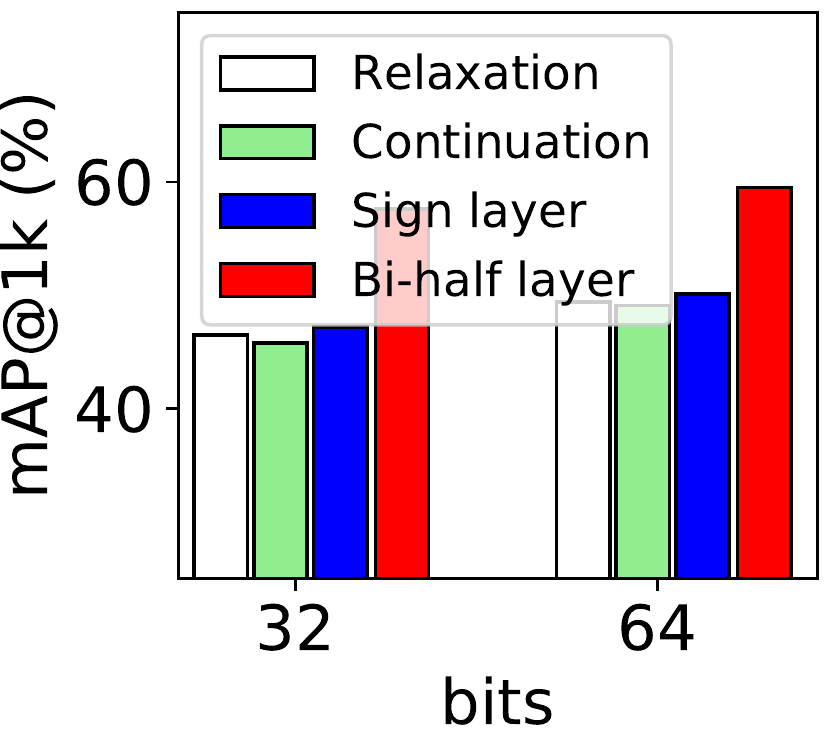}& \ \  \ \
\includegraphics[width=0.2065\textwidth]{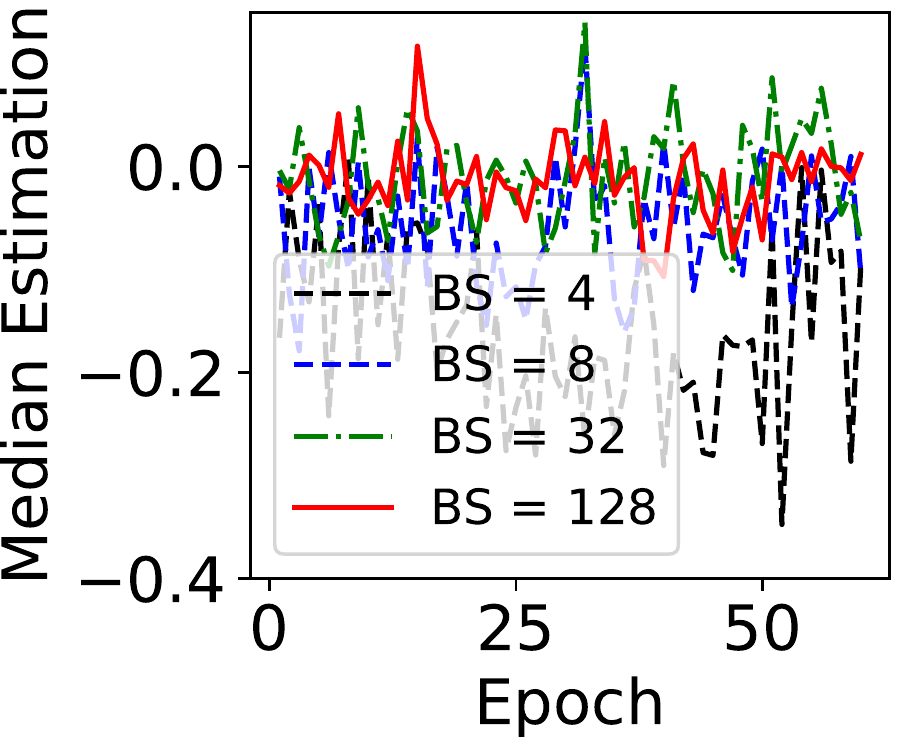}&
\includegraphics[width=0.195\textwidth]{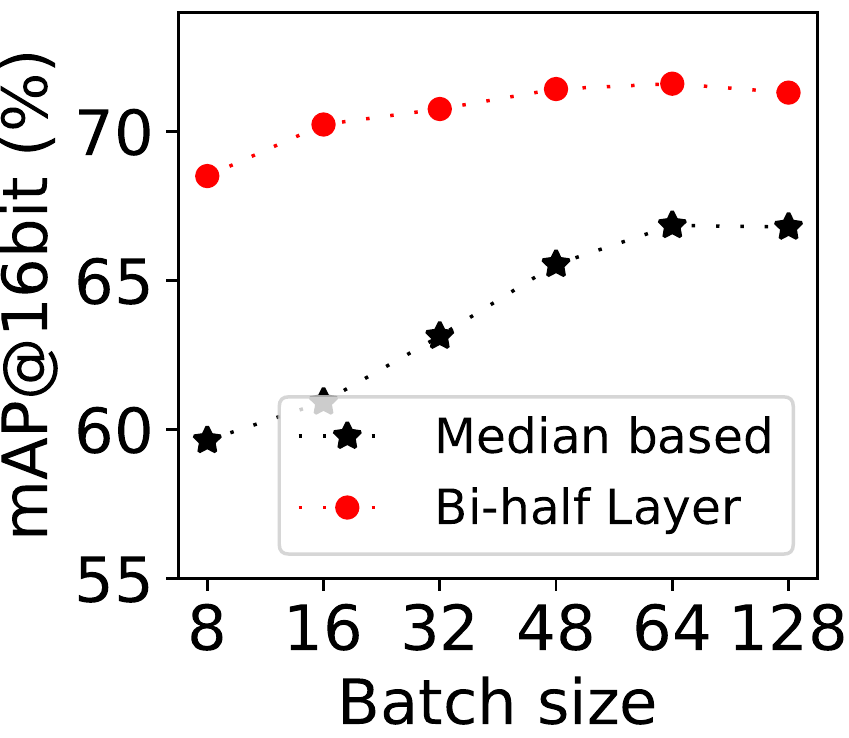}&
\includegraphics[width=0.191\textwidth]{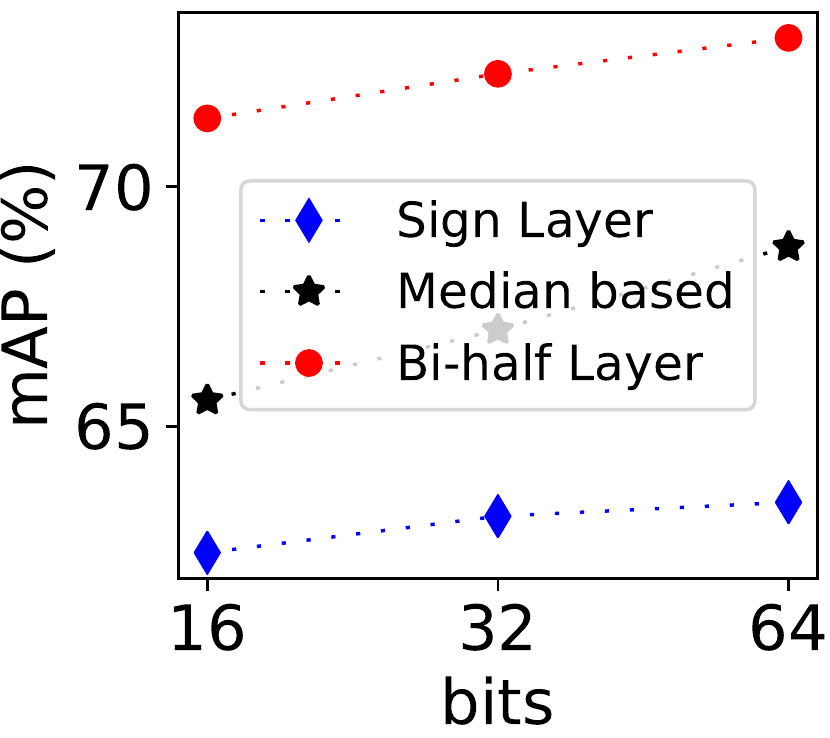} \\

   \multicolumn{5}{l}{  \ \ \ \  \ \ \ \ \ \ \ \
   (a)   left two
subfigures on Cifar-10
\ \ \ \ \ \ \  \ \ \ \ \ \ \ \ \ \ \ \ \  \ \ \ \ \ \ \ \ \ \ \ \ \ \ \ \ \ \ \ \  \ \ \ \ \ \ \ \
   (b) right three
subfigures on Flickr25k
   }
   \end{tabular}
\vspace{-0.1in}
\caption{  Empirical analysis  on Cifar-10 and Flickr25k datasets.
  (a) Our bi-half layer can generate informative hash bits and outperforms other coding methods; (b) the alternative median based method performs worse than bi-half layer.
 }
 \label{fig:ablation}
\end{figure*} \vspace{-0.1in}

\subsection{Training an AutoEncoder from scratch}
Our bi-half layer can embedded into current deep architectures to learn binary codes from scratch.
We train an AutoEncoder with a deep
encoder and decoder on Mnist datasets where encoder and decoder consist of two fc layers. We append our bi-half layer after encoder to generate binary code. The {reconstruction loss} is used as cost function. \textbf{We compare to using the sign layer and to adding an additional entropy regularization term in the loss}. For this baseline, as in~\cite{shen2018unsupervised,erin2015deep}, we use  $\mathbf{B}^\intercal \mathbf{1} $ as regularization term balanced with the \emph{BCE} reconstruction loss through a hyper-parameter $\alpha$.

%


In the top row of \fig{AE} we train the network with 3 bits and visualize the distributions of the continuous feature $\mathbf{U}$ over 5,000 images.
We observe that the features learned by sign layer are seriously tangled with each other.
With binarization, most images  will be scattered to same binary vertex  and thus some bits have no discriminative information.
By adding an entropy  regularization term, the feature tanglement can be mitigated,
 but it is suboptimal solution which requires careful hyper-parameter
tuning.
The proposed bi-half layer can learn evenly distributed features.

\fig{AE} (d) shows  the retrieval performance where the left two subfigures show the effect of hyper-parameters $\gamma$ in \eq{gradienet} and $\alpha$ of
 term  $\mathbf{B}^\intercal \mathbf{1} $.
 with code length 32 and the right one presents the mAP over different code lengths. Tuning the parameters can effectively improve the performance.
For  Sign+Reg method, it is a suboptimal solution in optimizing information entropy in comparison with bi-half layer, which can be
 further demonstrated in the right subfigure of \fig{AE} (d).
The reconstruction loss for sign layer and bi-half layer is shown in \fig{AE} (e).
 We observe that generating informative binary codes in latent space can effectively minimize the reconstruction loss.

\subsection{Empirical analysis}
%
For the pre-trained models, we follow the unsupervised setting in~\cite{NIPS_greedy_hash} and use  $|| cos(\mathbf{a}_1, \mathbf{a}_2) - cos(\mathbf{b}_1, \mathbf{b}_2) ||_2^2 $ as cost function to minimize the difference on the cosine distance
relationships, where $\mathbf{a}$ means the continuous feature extracted from the last layer of a pre-trained network of one sample while $\mathbf{b} $ means the corresponding binary code.

\begin{figure}[t] \vspace{-0.2in}
	\centering
	$\begin{tabular}{@{}c@{}c@{}}
	\includegraphics[width=0.49\linewidth]{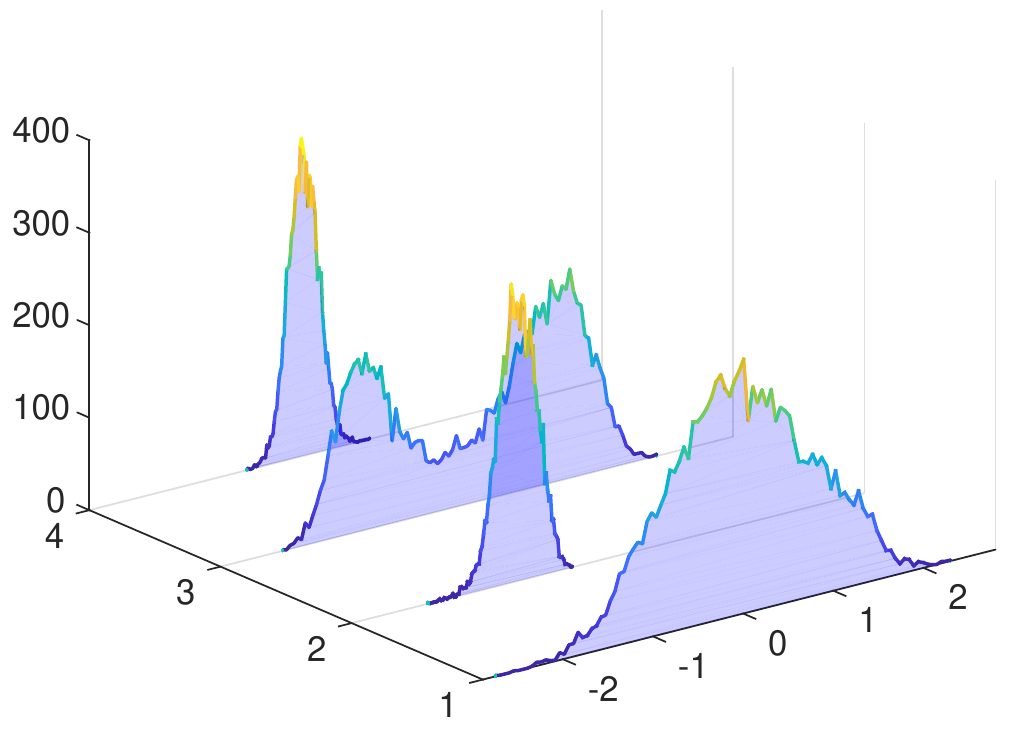} & \ \ \ \
	\includegraphics[width=0.49\linewidth]{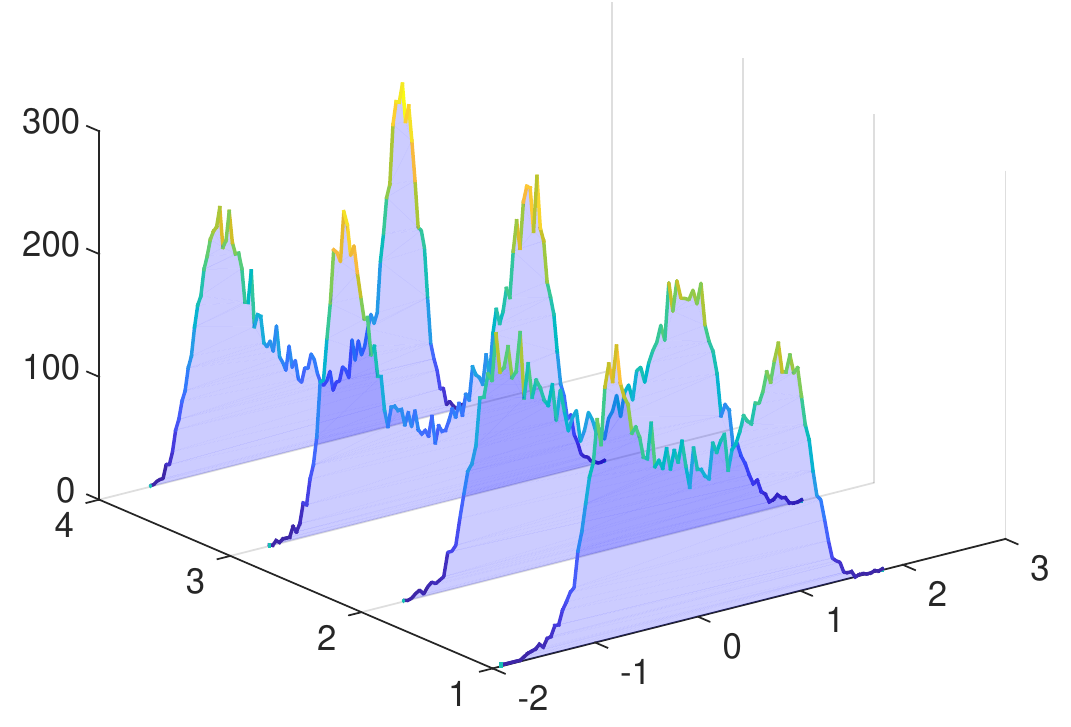} \\
	\small{Sign layer} \ \ \ \ &  \small{Bi-half layer}
	\end{tabular}$
\vspace{-0.12in}
	\caption{ Comparing the distribution of continuous feature $\mathbf{U}$ for training 4 bits with a sign layer (left) versus our bi-half layer (right) over all images in Cifar-10. The y-axis shows each of the 4 bit dimensions; the x-axis shows the continuous values in $\mathbf{U}$; the z-axis presents how many images contain such a continuous value (binned). In contrast to the sign layer, our bi-half method approximates the ideal half-half distribution.
	}
	\label{fig:diff_modules}
\end{figure}

\noindent \textbf{How are the continuous features distributed?}
In~\fig{diff_modules} we train the network on  Cifar-10(I) with 4 bits and visualize the histogram distributions of each dimension in the continuous encoded feature $\mathbf{U}$ over all images. The sign layer~\cite{cao2017hashnet,NIPS_greedy_hash} does not match an ideal half-half distribution   whereas our bi-half method does a better approximation.
%
%

\begin{table}[!t]
	\centering
\setlength{\tabcolsep}{1.7mm}{
	\renewcommand{\arraystretch}{1}
	\resizebox{1\linewidth}{!}{
		\begin{tabular}[c]{lccccccc}
			\toprule
			\multirow{2}{*}{Method} & \multicolumn{3}{c}{Cifar-10(I) } & & \multicolumn{3}{c}{Nus-wide(I)}\\
			& 16 bits & 32 bits  & 64 bits && 16 bits & 32 bits  & 64 bits   \\
			\midrule
			DeepBit  & 19.40 & 24.90 & 27.70  && 39.22 & 40.32 & 42.06    \\
			SAH & 41.80 & 45.60 & 47.40& & -- & -- & -- \\
            SADH  & -- & -- & --& & 60.14 & 57.99 & 56.33 \\
			HashGAN& 44.70 & 46.30 & 48.10 && -- & -- & --  \\
			GreedyHash$^\star$   & 44.80 & 47.20 & 50.10 & & 55.49 & 57.47 & 60.93 \\
			Ours & \textbf{56.10} & \textbf{57.60} &\textbf{59.50}& & \textbf{65.12} & \textbf{66.31} & \textbf{67.26} \\
			\bottomrule
		\end{tabular}
	} \vspace{-0.1in}
}	\caption{mAP@$1000$ results on Cifar-10(I) and mAP@All results on Nus-wide(I).  The $^\star$ denotes that we run the experiments with the released code by the authors. }
	\label{table:CIFAR10}	\vspace{-0.2in}
\end{table}

\noindent \textbf{How are individual hashing bits distributed?} In the left subfigure of \fig{ablation} (a)
we show the per-bit probability of code $+1$ over all images for 16 bits. Cifar-10(I) dataset is used to generate hash codes. The sign layer gives a non-uniformly distribution, and even for some bits the probability is completely zero or completely one: Those bits never change their value in the entire dataset and can thus safely be discarded. In contrast, our bi-half method approximates a uniform distribution, making good use of full bit capacity.

\noindent \textbf{Other hash coding strategies.}
The right subfigure of \fig{ablation} (a) shows the comparison between our bi-half coding method and three other hash coding strategies: continuous relaxation layer~\cite{CaoCVPR18,Liu2016DeepSH} ($\mathbf{B}\rightarrow \mathbf{U}$),  smoothed continuation layer~\cite{cao2017hashnet,LaiCVPR2015} ($\mathbf{B} \rightarrow \tanh(\beta \mathbf{U})$) and sign layer~\cite{NIPS_greedy_hash} ($\mathrm{sign}(\mathbf{U})$), respectively. Both $32$ and $64$ bits are used to generate hash codes on the Cifar-10(I) dataset.
From the results, we see that the sign layer
 method slightly outperforms the other two coding methods which is consistent with~\cite{NIPS_greedy_hash}.
This may be because the sign layer method can effectively keep the discrete constraint in comparison with other two methods.  Our bi-half method outperforms other methods for both code sizes.

\noindent \textbf{An alternative variant of bi-half layer.}
An alternative variant of the bi-half layer is to learn a translation term $t$ added to the sign function $\mathrm{sign}(\mathbf{u} +t)$ for each hash bit to get half-half distributed binary codes.
We estimate the median statistic over mini-batches to implement this idea.
Specifically, we keep an exponential moving average (EMA) of median points over each mini-batch which is used during inference.
We conduct the comparison on Flickr25k dataset in \fig{ablation} (b).
The left subfigure of \fig{ablation} (b)  shows how the EMA estimation of median changes with
the training epochs over different batch sizes.
We adopt the linear learning rate scaling rule~\cite{goyal2017accurate,krizhevsky2014one} to adapt to batch size. We note that smaller batch size makes the  estimation value unstable. The middle subfigure conducts a comparison between bi-half layer method and median translation method  over different batch sizes on using 16  bits.
Increasing batch size can significantly improve the performance for median based method.
Due to memory limitations, unfortunately, it is difficult to use very large batch sizes.
The left  subfigure  of \fig{ablation} (b) shows the comparison with greedy hash (sign layer) and the median-based method with code length 16, 32 and 64. As expected, adding median term increases the sign layer baseline and bi-half layer significantly outperforms median-based approach.

 \vspace{-0.1in}

%

\begin{figure*}[t]\vspace{-0.15in}
\centering
\begin{tabular}{llllllll}
\includegraphics[width=0.2\textwidth]{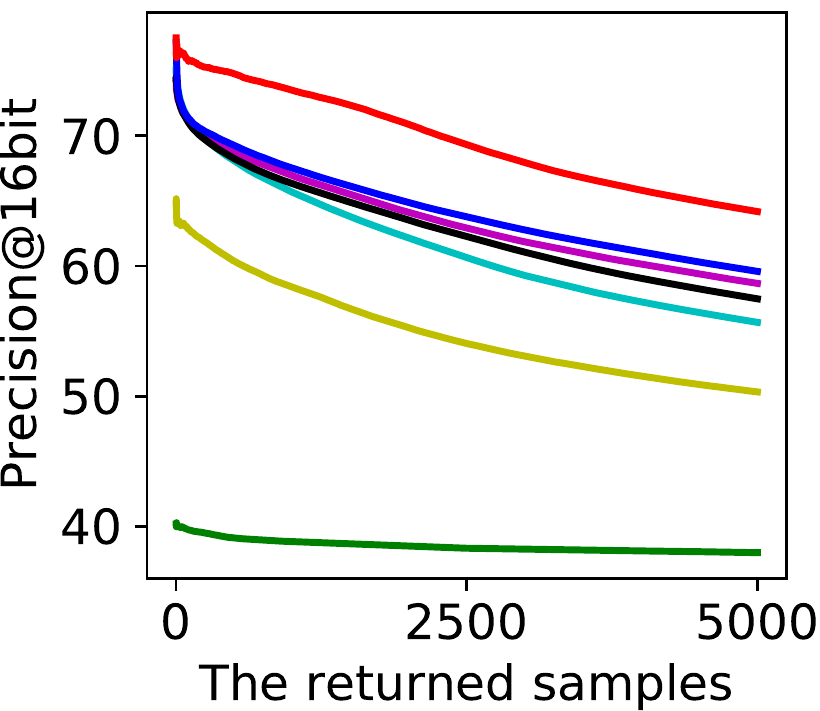} &
\includegraphics[width=0.2\textwidth]{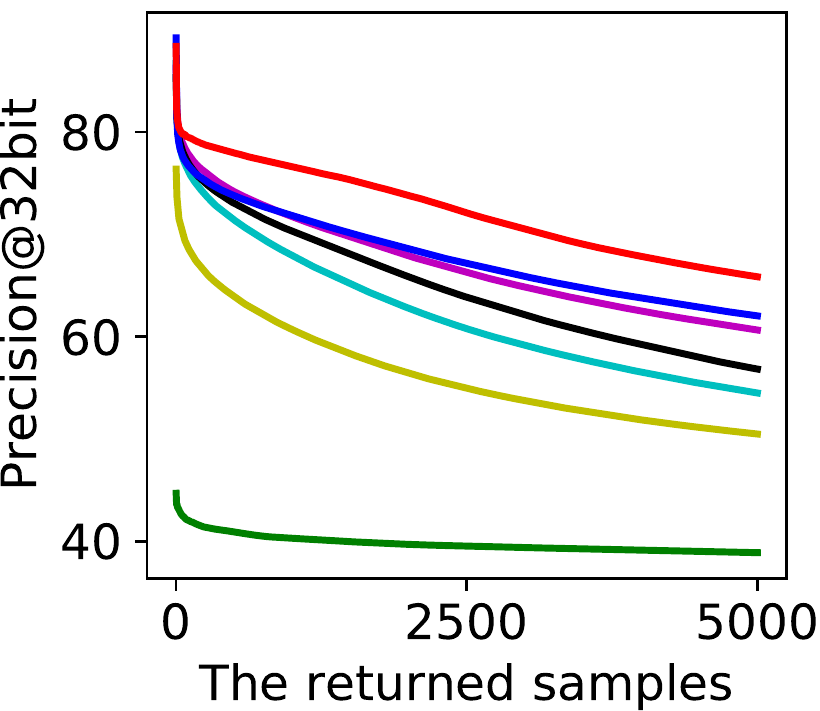} &
\includegraphics[width=0.197\textwidth]{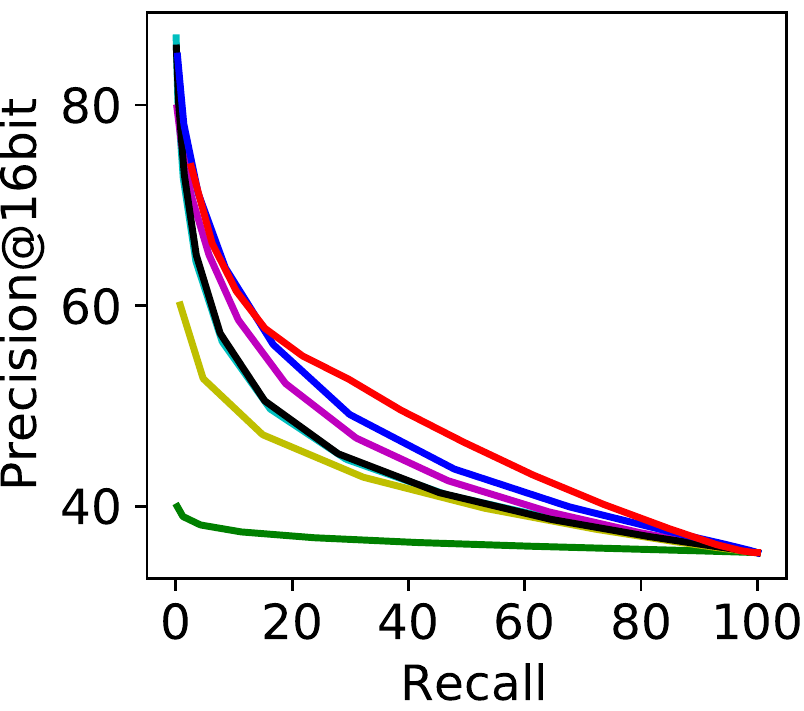}&
\includegraphics[width=0.299\textwidth]{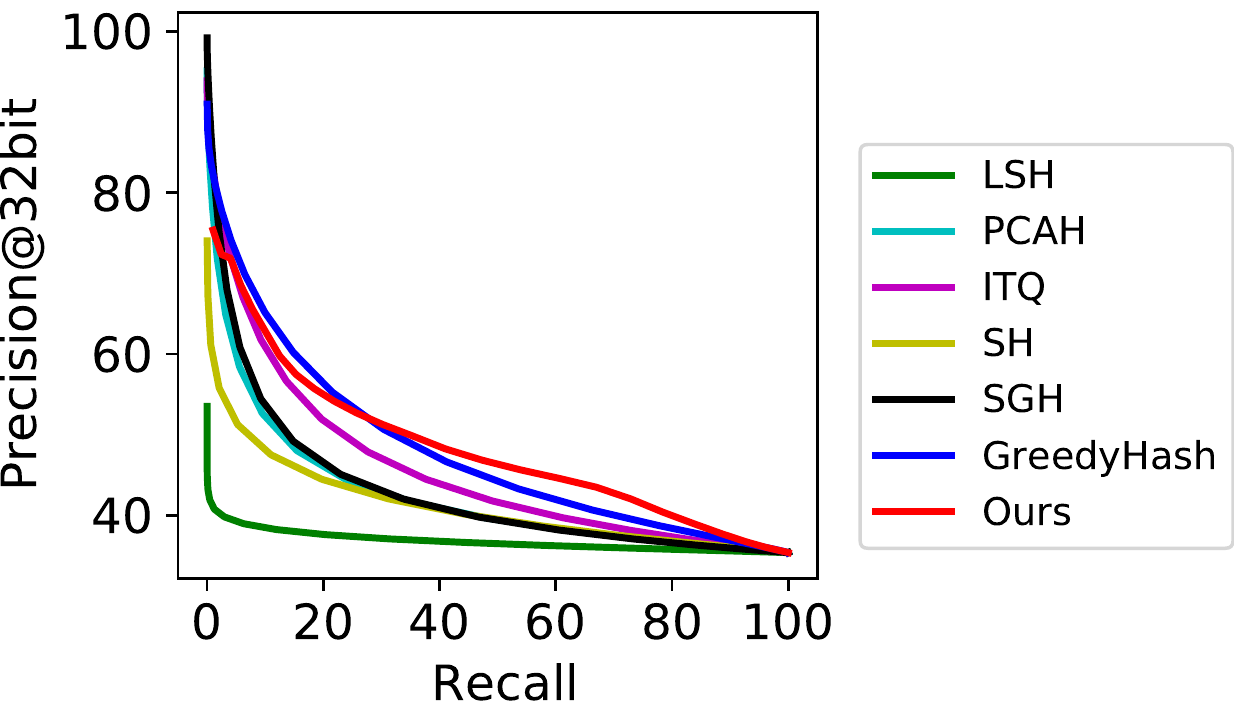}
   \end{tabular} \vspace{-0.1in}

\caption{  Top N precision and precision-recall curves on Mscoco.  The proposed bi-half layer performs best. 
   }
 \label{fig:prcurvesmscoco}
\end{figure*}

%
%

\begin{table*}[!t]
	\centering
	\footnotesize
 \setlength{\tabcolsep}{1.65mm}{
		\begin{tabular}[c]{lccccccccccc}
			\toprule
			\multirow{2}*{Method} & \multicolumn{3}{c}{Flickr25k} & & \multicolumn{3}{c}{Nus-wide(II)}  &  & \multicolumn{3}{c}{Cifar-10(II)}\\
			& 16 bits & 32 bits  & 64 bits  & & 16 bits   &  32 bits & 64 bits & & 16 bits & 32 bits &  64 bits \\
			\midrule
			LSH + VGG~\cite{Datar04localitysensitivehashing} &{58.31}&{ 58.85 } & 59.33& & {43.24} & {44.11} &{44.33} & &{13.19}&{ 15.80} &{16.73} \\
			SH + VGG~\cite{Weiss2008Spectral} &{59.19}&{ 59.23 } & 60.16& & {44.58} & {45.37} &{49.26} & &{16.05}&{ 15.83} &{15.09} \\
			ITQ + VGG~\cite{Gong11iterativequantization} &{61.92}&{ 63.18 } & 63.46& & {52.83} & {53.23} &{53.19} & &{19.42}&{ 20.86} &{21.51} \\
			DeepBit~\cite{lin2016learning} &{59.34}&{ 59.33 } & 61.99& & {45.42} & {46.25} &{47.62} & &{22.04}&{ 24.10} &{25.21} \\
			SGH~\cite{dai2017stochastic} &{61.62}&{ 62.83 } & 62.53& & {49.36} & {48.29} &{48.65} & &{17.95}&{ 18.27} &{18.89} \\
			SSDH~\cite{yang2018semantic} &{66.21}&{ 67.33 } & 67.32& & {62.31} & {62.94} &{63.21} & &{25.68}&{ 25.60} &{25.87} \\
			DistillHash~\cite{yang2019distillhash} &{69.64}&{ 70.56 } & 70.75& & {66.67} & {67.52} &{67.69} & &{28.44}&{ 28.53} &{28.67} \\
			GreedyHash$^\star$~\cite{NIPS_greedy_hash} &{62.36}&{ 63.12 } & 63.41& & {51.39} & {55.80} &{59.27} & &{28.71}&{31.72} &{35.47} \\
			Ours &\textbf{71.42}&\textbf{ 72.35} & \textbf{73.10}& & \textbf{67.12} & \textbf{68.05} &\textbf{68.21} & &\textbf{42.87}&\textbf{43.29} &\textbf{44.13} \\
			
			\bottomrule
		\end{tabular}
	}
	\vspace{-0.1in}
	\caption{mAP@All for various methods on three Flickr25k, Nus-wide(II) and  Cifar-10(II) datasets. Our method with 16 bits outperforms others that use 64 bits.}
	\label{table:experiment_threedata}\vspace{-0.1in}
\end{table*}

\subsection{Comparison with state-of-the art}
We compare our method with previous unsupervised hashing methods, including seven shallow
unsupervised hashing methods, \emph{i.e.} LSH~\cite{Datar04localitysensitivehashing}, SH~\cite{Weiss2008Spectral},
PCAH,
ITQ~\cite{Gong11iterativequantization},
 SGH~\cite{Jiang2015Scalable},
  and eight deep unsupervised hashing
methods, \emph{i.e.} DeepBit~\cite{lin2016learning}, SGH~\cite{dai2017stochastic}, SSDH~\cite{yang2018semantic},
DistillHash~\cite{yang2019distillhash}, SAH~\cite{do2017simultaneous}, HashGAN~\cite{ghasedi2018unsupervised}, SADH~\cite{shen2018unsupervised}, and  GreedyHash~\cite{NIPS_greedy_hash}.
To have a fair comparison, we adopt the deep features for all shallow
architecture-based baseline methods.
For GreedyHash, we run the experiments with
the released code by the authors.
 For other methods, the results are taken from the related literatures.
%

\tabl{CIFAR10} shows the   mAP@$1000$ results on Cifar-10(I) and mAP@All results on Nus-wide(I)
over three different hash code sizes 16, 32 and 64.
The compared greedy hash~\cite{NIPS_greedy_hash} method which directly uses the sign function as hash coding layer outperforms  everything except our method for all code sizes. Greedy hash~\cite{NIPS_greedy_hash} is effective to solve the vanishing gradient problem and maintain the discrete constraint in hash learning, but it cannot maximize hash bit capacity. In contrast, our method does maximize hash bit capacity and clearly outperforms all other methods on this two datasets.

In \tabl{experiment_threedata}, we present the mAP results on three datasets Flickr25k, Nuswide(II), and Cifar-10(II), with hash code length varying from 16 to 64. The experiments are conducted with the same setting as in the compared methods. We do best for all hash bits sizes for all three datasets.

In
\fig{prcurvesmscoco}, we conduct experiments on more challenging Mscoco dataset.
The left two subfigures  present
the TopN-precision curves with code lengths 16 and 32. Consistent
with mAP results, we can observe that our method performs best.
Both mAP and TopN-precision curves are Hamming ranking based metrics where our method can achieve superior performance.
Moreover,  we plot the precision-recall curves for
all methods with hash bit lengths of 16 and 32 in the right
two subfigures  \fig{prcurvesmscoco} to illustrate the
hash lookup results. From the results, we
can again observe that our method consistently achieves
the best results among all approaches, which further demonstrates the superiority of our proposed method.


Hashing is about  compact storage and fast retrieval, thus we analyze using fewer bits in \tabl{CIFAR10},  \tabl{experiment_threedata}  and \fig{prcurvesmscoco}. Only for Nus-wide(II) we perform on par while in all other datasets our method using 16 bits clearly outperforms other methods using 64 bits. This shows a 3 times reduction in storage and speed while even improving accuracy.

\begin{table}[!t]
	\centering
	\renewcommand{\arraystretch}{1.2}
	\resizebox{1\linewidth}{!}{
\setlength{\tabcolsep}{0.43mm}{
		\begin{tabular}[c]{llccccccc}
			\toprule
			  \multirow{2}*{Backbone} & \multirow{2}*{Method}& \multicolumn{3}{c}{Ucf-101} & &  \multicolumn{3}{c}{Hmdb-51}\\
			& &16 bits &  32 bits & 64 bits & & 16 bits &  32 bits & 64 bits   \\ \midrule		
		\multirow{2}*{ResNet-34} 	 & GreedyHash$^\star$ & 45.49 & 57.24 & 64.77  & & 30.32 & 37.55& 40.53\\
			    & Ours  & 50.83 & 60.30& 65.89& & 34.21 &  38.67 & 41.74 \\  \midrule		
			\multirow{2}*{ResNet-101} & GreedyHash$^\star$  & 39.29 & 58.35 & 67.23  & & 27.60 & 39.96& 42.07  \\
			     & Ours & \textbf{59.30} & \textbf{66.13} & \textbf{68.47} & & \textbf{36.68} & \textbf{41.48}  & \textbf{43.03}  \\
			\bottomrule
		\end{tabular}
	}}\vspace{-0.1in}
	\caption{mAP@100 results on two video datasets using kinetics pre-trained 3D ResNet-34 and 3D ResNet-101. The $^\star$ denotes we run the experiments with the released code. }
	\label{table:video}
	\vspace{-0.15in}
\end{table}

\vspace{-0.02in}\textbf{Video Retrieval Results:}
In \tabl{video}, we present the mAP@100 results for Ucf-101 and Hmdb-51 datasets with code length 16, 32 and 64.
For both datasets and both ResNet models our bi-half method consistently outperforms the sign layer method~\cite{NIPS_greedy_hash} over all hash bit length, especially for short bits.  In hashing, fewer bits is essential to save storage and compute.

\section{Conclusion}  
We propose a new parameter-free Bi-half Net for unsupervised hashing learning by optimizing bit entropy. Our Bi-half layer has no hyper-parameters and compares favorably to minimizing bit entropy with an additional hyper-parameter in the loss.
The designed bi-half layer can be easily embedded into current deep architectures, such as AutoEncoders, to automatically generate higher quality binary codes.
The proposed proxy derivative in backward pass can effectively encourage the continuous feature
distribution to align with the ideal half-half distributed distribution. One limitation is that the independence between different bits is not considered, which will be investigated in future work.
Experiments on 7 datasets show state of the art results. We often outperform other hashing methods that use 64 bits where we need only 16 bits.

\clearpage

	{
		\bibliographystyle{aaai}
		\bibliography{egbib}
	}

%
%

\end{document}